\documentclass{article}

\PassOptionsToPackage{square,numbers}{natbib}
\usepackage[sglblindworkshop, final]{neurips_2025}

\usepackage[utf8]{inputenc} 
\usepackage[T1]{fontenc}    
\usepackage{hyperref}       
\usepackage{url}            
\usepackage{booktabs}       
\usepackage{amsfonts}       
\usepackage{nicefrac}       
\usepackage{microtype}      
\usepackage{xcolor}         
\usepackage{graphicx}
\usepackage{amsmath}
\usepackage{wrapfig}
\usepackage{subcaption}
\usepackage{makecell}

\title{Bayesian Surrogates for Risk-Aware Pre-Assessment of Aging Bridge Portfolios}
\workshoptitle{MLxOR: Mathematical Foundations and Operational Integration of Machine Learning for Uncertainty-Aware Decision-Making}

\author{
Sophia V. Kuhn$^{1,3,*}$ \quad Rafael Bischof$^{2,3}$ \quad Marius Weber$^{1}$ \quad Antoine Binggeli$^{1}$\\
\textbf{Michael A. Kraus}$^{4}$ \quad \textbf{Walter Kaufmann}$^{1,3}$ \quad \textbf{Fernando Pérez-Cruz}$^{4,5}$ \\ \\
$^{1}$Institute of Structural Engineering, ETH Zurich, Switzerland\\
$^{2}$Computational Design Lab, ETH Zurich, Switzerland\\
$^{3}$Design ++, ETH Zurich, Switzerland\\
$^{4}$Institute of Structural Mechanics and Design, TU Darmstadt, Germany\\
$^{5}$Department of Computer Science, ETH Zurich, Switzerland\\
$^{6}$Bank for International Settlements, Switzerland\\
*Correspondence to \texttt{kuhnso@ethz.ch}
}

\begin{document}

\maketitle

\begin{abstract}
Aging infrastructure portfolios pose a critical resource allocation challenge: deciding which structures require intervention and which can safely remain in service. 
Today’s structural assessment approaches do not scale to portfolio level, as they require time-consuming manual digital modeling and computationally expensive simulations for each structure.
We propose Bayesian neural network (BNN) surrogates for rapid structural pre-assessment of worldwide common bridge types, such as reinforced concrete frame bridges. Trained on a large-scale database of non-linear finite element analyses generated via a parametric pipeline and developed based on the Swiss Federal Railway's bridge portfolio, the models accurately and efficiently estimate high-fidelity structural analysis results by predicting code compliance factors with calibrated epistemic uncertainty. 
Our BNN surrogate enables fast, uncertainty-aware triage: flagging likely critical structures and providing guidance where refined analysis is pertinent. 
We demonstrate the framework's effectiveness in a real-world case study of a railway underpass, showing its potential to significantly reduce costs and emissions by avoiding unnecessary analyses and physical interventions across entire infrastructure portfolios.
\end{abstract}





\section{Introduction}
Bridges are critical components of transportation infrastructure, facilitating connectivity and economic activity. 
Yet, many of these structures around the world are aging, with a large share now operating close to or beyond their intended service life \cite{betonkalender2010}. 
This presents asset managers with a difficult and resource intensive challenge: deciding which structures require immediate intervention, and how extensive those interventions should be. Accurate structural analysis is central to this process to determine whether a bridge meets safety requirements or must be strengthened or replaced.

Current assessment practices follow a tiered, so called levels-of-approximation approach \cite{fibMC2020}: 
Starting with simplified, conservative analyses that contain large safety margins to compensate for modeling abstractions. 
If these methods do not demonstrate compliance suggesting that an intervention is necessary, asset managers can commission more refined simulations such as non-linear finite element analysis (NLFEA). 
While NLFEA is expensive, time-consuming, and requires detailed inputs and expertise, 
it can uncover structural reserve capacity, avoiding costly strengthening measures and enabling structures to remain in service beyond their original design life.
However, the benefit of such detailed analyses is unknown a priori. In some cases, NLFEA merely confirms the findings of conservative methods at a significant extra cost. 
Furthermore, the typically high manual modeling and computational effort involved makes current assessment workflows unscalable to portfolio level.
%

In this work \cite{Kuhn2025}, we develop data-driven surrogate models that efficiently estimate refined structural analysis results and quantify the uncertainty associated with each prediction. 
These uncertainty-aware outputs enable risk-informed decisions that help prioritize allocation of scarce economic, material, and labor resources where they are most needed.
Our main contributions are 
(i) a parametric simulation pipeline that generates a large-scale, high-fidelity dataset of reinforced concrete frame bridges covering common structures of existing infrastructure portfolios;  
(ii) Bayesian Neural Network (BNN) \cite{pmlr-v37-hernandez-lobatoc15} surrogates trained to predict code compliance factors expected from detailed analyses, with calibrated epistemic uncertainty; and 
(iii) a triage policy that leverages this predictive distribution to inform decision-making at the portfolio level. 

Data-driven surrogates have been widely applied to approximate computationally expensive simulations in engineering, including structural health monitoring and capacity estimation \cite{BALMER_KUHN2024_CVAE, Kraus2024Glass_StrengthLabAI,NIPS2017_DeepEnsebles, GPR_williams_1995}. However, for structural engineering most approaches remain deterministic and lack uncertainty quantification. Probabilistic methods such as Gaussian Processes, deep ensembles, and BNNs provide predictive distributions instead of point estimates \cite{pmlr-v37-hernandez-lobatoc15}, with recent studies demonstrating their effectiveness for pavement response prediction \cite{BNNs_longSpanBridgesVibrations_2012}, bridge integrity assessment \cite{BNNs_pavementFEM_2022}, and materials modeling \cite{BNNs_compositeMaterialsFEM_2021}.
\begin{figure}[ht]
  \centering
  \includegraphics[width=\linewidth, trim=0 0 20 0, clip]{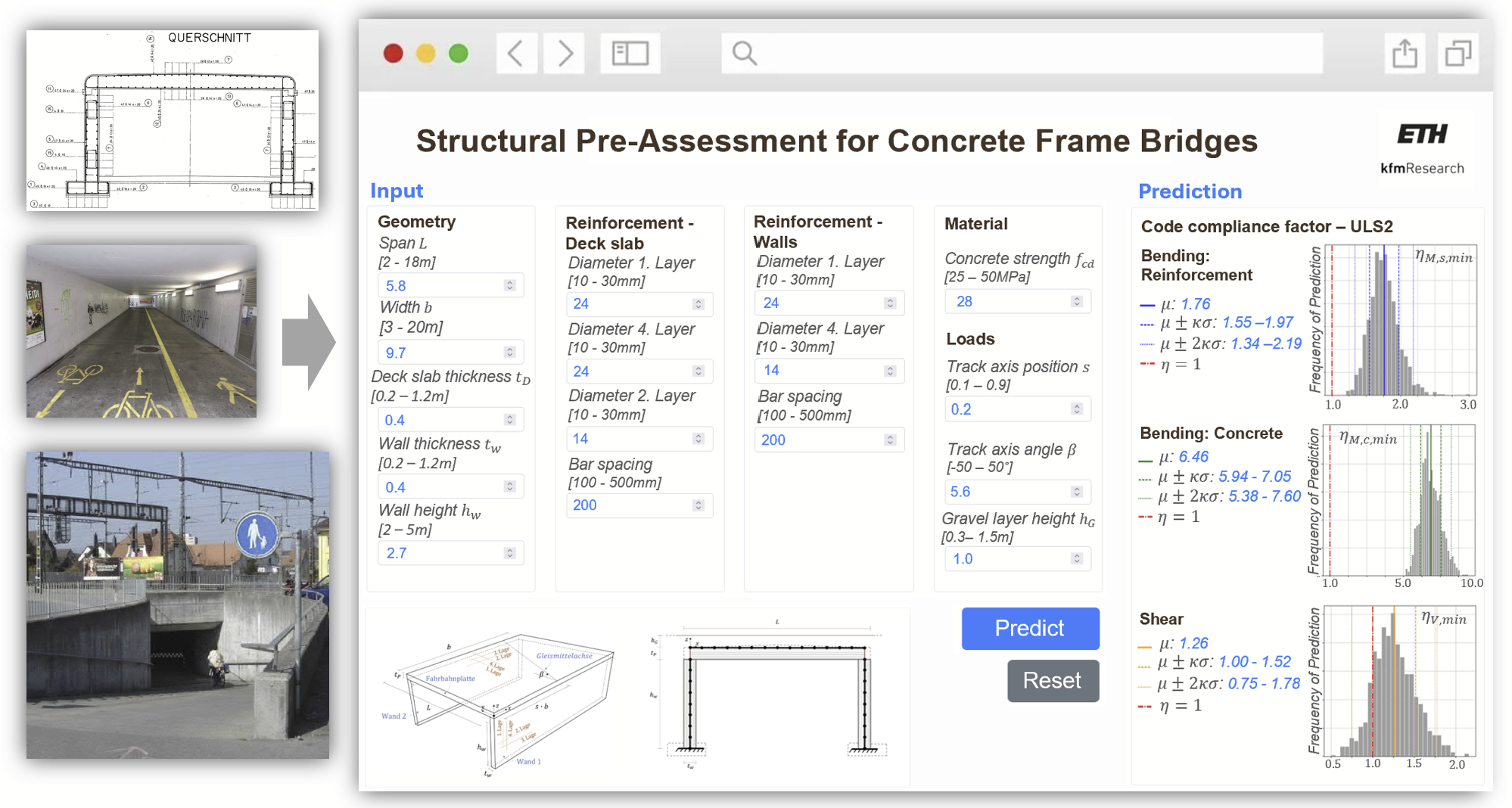}
  \caption{
    Graphical user interface of the developed ML-based structural pre-assessment tool for reinforced concrete frame bridges. 
    Users input structural parameters (left) and receive rapid code compliance predictions with uncertainty estimates (right). 
  }
  \label{fig:UI}
\end{figure}
\begin{table}[ht]
  \centering
  \caption{Decision policy for classifying structures based on surrogate predictions.}
  \label{tab:triage_policy}
  \begin{tabular}{@{}clp{10cm}@{}}
    \toprule
    \textbf{Class} & \textbf{Condition} & \textbf{Interpretation} \\
    \midrule
    {\color{red}\textbullet} Red &
    $\mu < 1$ &
    Critical. Prioritize this structure for immediate detailed analysis. Even refined methods may not demonstrate sufficient load-bearing capacity. \\
    \addlinespace[1ex]
    {\color{orange}\textbullet} Orange &
    \makecell[l]{$\mu > 1$ and \\ $\mu - 2\kappa\sigma < 1$}
     &
    Likely safe, but uncertainty is high. A refined analysis is recommended, especially if the predicted compliance factor is only slightly above 1. \\
    \addlinespace[1ex]
    {\color{green!50!black}\textbullet} Green &
    \makecell[l]{$\mu > 1$ and \\ $\mu - 2\kappa\sigma \geq 1$} &
    High confidence in compliance. Not to prioritize. Simplified analysis methods are predicted to be sufficient. \\
    \bottomrule
  \end{tabular}
\end{table}
\section{Methods}
%
Let $\mathbf{x}\in\mathcal{D}\subset\mathbb{R}^k$ denote a set of bridge parameters (geometry, materials, loading, etc.). 
We define a parametric high-fidelity mechanical analysis model $f(\mathbf{x})$ that returns code-compliance factors 
$\boldsymbol{\eta}(\mathbf{x})$, defined as resistance-to-demand ratios evaluated for structural safety code verifications. 
Since evaluating $f$ via materially non-linear finite element analysis (NLFEA) is computationally intensive, we approximate it using a Bayesian Neural Networks (BNNs) trained on a limited number of simulations. 
The BNN surrogate yields a predictive distribution $p(\hat y\mid \mathbf{x},\mathcal{D})$; its mean $\mu$ serves as the best estimate of $\eta$, and its standard deviation $\sigma$ quantifies epistemic uncertainty from limited data / model capacity. 
Decisions follow the predictive distribution, 
mapping to the triage policy in Table~\ref{tab:triage_policy}.

\textbf{Data generation with parametric NLFEA.}
To obtain training data for supervised learning, we built an automated pipeline that generates models of reinforced concrete frame bridges from a parameter vector, evaluates them with NLFEA, and extracts code-compliance labels. 
The pipeline links the programs Rhino/Grasshopper (for geometry generation) with Ansys Mechanical APDL (for NLFEA), where the Cracked Membrane Model \cite{kaufmann1998cracked,TRW_CMM} is implemented as a layered shell element user material. This orchestration is automated via the StrucEngLib plug-in \cite{StrucEngLib}. 
For each simulation, the pipeline validates structural safety according to Swiss design codes and outputs three global compliance factors $\boldsymbol{\eta}(\mathbf{x})=$$(\eta_{M,c,\min},\eta_{M,s,\min},\eta_{V,\min})$. These quantify the structural safety margin in bending (for concrete and reinforcing steel) and shear, computed as the minimum resistance-to-demand ratios across all elements, with $\eta \ge 1$ indicating code compliance. 
Since each NLFEA run is expensive ($\approx\!25$ minutes per bridge, $\approx\!2.5$ minutes with 10-way parallelism)\footnote{Note that in practice, the high cost of NLFEA is caused less by the simulation time itself and more by the manual effort required to build a validated digital model of a specific bridge.}, we adopt a sampling strategy that balances coverage and efficiency. 
To ensure realistic variability, parameter ranges were derived from a database analysis of the Swiss Federal Railways (SBB) bridge inventory, with expert feedback.
Latin Hypercube Sampling (LHS) is used to broadly cover the input space, followed by adaptive resampling based on kernel density estimation to focus simulations near the safety-critical region around $\eta \approx 1$ (cf.\ Table~3 in \cite{Kuhn2025}). 
The resulting dataset of $\approx\!11$k samples spans common structures in the SBB portfolio while providing high resolution near the decision boundary.

\textbf{Bayesian surrogate.}
We train three BNNs $\mathcal{B}_\theta$, one per compliance factor, 
adopting stochastic variational inference, approximating the true posterior as a product of independent Gaussian distributions for each model parameter $\theta$ following the mean-field approximation \cite{murphy2012machine} and using Gaussian priors $p(\theta)=\mathcal{N}(0,\,0.1I)$.
The objective function, derived by maximizing the evidence lower bound (ELBO) \cite{blundell2015weightuncertaintyNN}, combines a KL divergence regularization term with an enhanced, domain-adapted weighted MSLE:
\begin{equation}
L(\theta,D) = \lambda \cdot \text{KL}(q_\phi(\theta) \| p(\theta)) + \text{wMSLE}(y,\mu(\hat{y}))
\label{eq:LossFunction}
\end{equation}
where $D$ is a set of training samples, $y$ is the true value, and $\mu_i(\hat{y})$ is the mean predicted value.
The wMSLE heavily penalizes $\eta$-values $\approx 1$,  focusing learning near the decision boundary.
The predictive mean $\mu$ and uncertainty $\sigma$ are estimated using 20 forward passes during training and 1000 at inference. 
In addition to regularization during training, to further improve alignment between predicted and observed uncertainties, we apply post-hoc rescaling of the predictive standard deviations by a constant factor $\kappa$.
%
Finally, to enable deployment when detailed structural information is unavailable, we use Kernel SHAP \cite{NIPS2017_7062} to rank input features by importance and perform reduced-input inference by sampling missing features with LHS and propagating them through the trained BNNs.

\section{Results}
Model accuracy is evaluated on held-out data using RMSE and MAPE globally and stratified across $\eta$ ranges to evaluate performance in regions with different importance for decision making. Calibration of the uncertainty quantification is assessed with calibration curves and summary metrics (Total Calibration Error (TCE) and Calibration Bias (CB) \cite{Kuhn2025}). 

\begin{figure}[t]
  \centering
  \begin{minipage}[c]{0.45\linewidth}
    \centering
    \small
    \begin{tabular}{lccc}
      \toprule
      \textbf{Target} & \textbf{RMSE} & \textbf{MAPE} & \textbf{CB} \\
      \midrule
      $\eta_{M,s,\min}$ & 0.10 & 4.8\% & $-0.8$ \\
      $\eta_{M,c,\min}$ & 0.40 & 37.5\% & $-0.9$ \\
      $\eta_{V,\min}$   & 0.60 & 46.7\% & $-0.6$ \\
      \bottomrule
    \end{tabular}
    \captionof{table}{Predictive performance in the safety-critical region
    $\eta \in [0.5,1.5]$. CB reported after post-hoc scaling.}
    \label{tab:perf_bnn_critical}
  \end{minipage}%
  \hfill
  \begin{minipage}[c]{0.5\linewidth}
    \centering
    \includegraphics[width=0.65\linewidth]{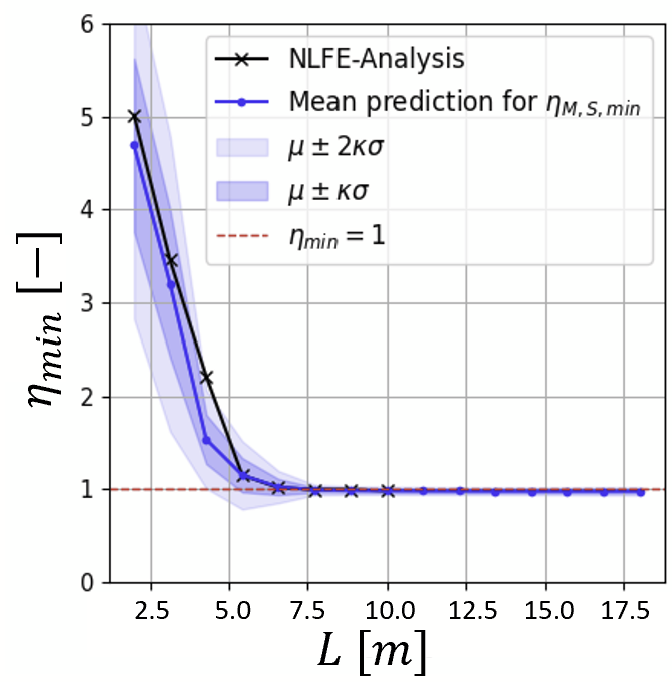}
    \caption{Representative 1D parameter study example comparing BNN predictions with NLFEA.}
    \label{fig:parameter-studies}
  \end{minipage}
\end{figure}

\textbf{Decision-focused accuracy.}  
Prediction accuracy and uncertainty estimation are strongest in the safety-critical range $0.5 \leq \eta \leq 1.5$ (Fig.~\ref{fig:parameter-studies}). 
Predictive intervals narrow near $\eta \approx 1$ and widen in uncritical regions, due to our targeted resampling and dedicated loss.  
Among the targets, $\eta_{M,s,\min}$ is predicted most reliably (RMSE 0.10, MAPE 4.8\%), $\eta_{M,c,\min}$ shows moderate error, and $\eta_{V,\min}$ remains most challenging due to the fact that shear is governed by abrupt local effects (cf.\ Table~\ref{tab:perf_bnn_critical}).

\textbf{Calibration of predictive uncertainty.}  
After training, the predictive intervals computed as $\mu \pm z(p)\sigma$, where $z(p)$ denotes the standard normal quantile for a given confidence level $p$, indicate that the BNN outputs are mildly overconfident
in their uncertainty estimates.
These intervals capture fewer true values than expected, causing calibration curves to fall below the diagonal and CB to be positive. We therefore apply a post-hoc scale factor per head (chosen on the validation set: $\kappa=1.3$ for steel, $1.7$ for concrete, $2.8$ for shear) to align nominal and empirical coverage, reduce TCE, and obtain mildly under-confident intervals (CB $< 0$, Table \ref{tab:perf_bnn_critical}) on the held-out data, which is desirable in this safety-critical setting. 
The Gaussian predictive distribution assumption holds well overall, but less so for shear, due to its brittle and locally governed behavior.

\textbf{Reduced-input deployment.}  
Using only the top five most influential and readily available input features (i.e., span,  plate thicknesses, height and width), the reduced-input variant retains most of the predictive accuracy for steel and shear (predominantly governed by geometry) while accuracy for concrete degrades more noticeably, consistent with the SHAP analysis. 
Uncertainty calibration remains well-aligned, supporting the use of the reduced-input model for early screening, with optional input refinement when higher prediction resolution is required.

\textbf{Case study: railway underpass.}
We demonstrate the surrogate on a Swiss railway underpass using the deployed graphical interface (Fig.~\ref{fig:UI}). 
The model can be queried at multiple levels of input detail: from basic geometric parameters 
to detailed reinforcement and loading. 
SHAP-based feature attribution helps prioritize which inputs to acquire next to most effectively reduce uncertainty, enabling progressive refinement.
At all input levels, the prediction intervals $\mu \!\pm\! 2\kappa\sigma$ cover the NLFEA ground truth, confirming robust calibration.
Already with only high-level inputs, the model flags shear as potentially critical, though with high uncertainty.
As detailed input is provided, intervals narrow. At full detail, the BNN confidently predicts compliance in bending and identifies shear as the decisive verification (shown in Fig. \ref{fig:UI}). 
The triage policy (Table~\ref{tab:triage_policy}) classifies this example structure as  orange, and recommends its refined analysis.
Subsequent NLFE analysis confirmed compliance and revealed reserve capacity, declaring interventions therefore as unnecessary. In contrast, simplified analysis would have incorrectly triggered strengthening of the structure.
For this single structure alone, avoiding such unnecessary intervention saved approximately \$~370,000 (estimate by SBB AG), as well as substantial material use, CO\textsubscript{2} emissions, and traffic disruptions on and under the overpass, which demonstrates our tool’s economic and environmental value for portfolio-level decision-making.

\section{Conclusion}

We present a scalable, uncertainty-aware surrogate for structural pre-assessment of common bridge types. 
Trained on high-fidelity NLFEA data, our BNN model efficiently predicts code compliance based on a few structural input parameters with sufficient accuracy and calibrated uncertainty, supporting crucial triage decisions and whether refined analysis is warranted.
Demonstrated on a real-world case study, the approach enables portfolio-scale screening even from limited input detail. It helps reduce unnecessary assessments and focuses scarce resources where they are most impactful, a critical need given today’s aging infrastructure portfolios.

\begin{ack}
We acknowledge support and data access from Swiss Federal Railways (SBB). See journal publication \cite{Kuhn2025} for extended acknowledgments.
\end{ack}


\medskip

\bibliography{references}
\bibliographystyle{plainnat}


\end{document}